\documentclass[10pt,twocolumn,letterpaper]{article}
\usepackage{cvpr}
\usepackage{times}
\usepackage{epsfig}
\usepackage{graphicx}
\usepackage{amsmath}
\usepackage{amssymb}

\usepackage{empheq}
\usepackage{hyperref}
\usepackage[switch, pagewise]{lineno}
\usepackage{xcolor}
\usepackage{units}
\usepackage{graphicx}
\usepackage{amssymb}
\usepackage{graphicx}
\usepackage{epstopdf}
\usepackage{subfig}
\usepackage{array}
\usepackage{amsmath}
\usepackage{url}
\usepackage{pdflscape}
\usepackage{rotating,booktabs}
\usepackage{multirow}
 \usepackage{gensymb}
\usepackage{caption}
\usepackage{siunitx} 
\usepackage{esvect}

\usepackage{mathtools}

\DeclarePairedDelimiter\floor{\lfloor}{\rfloor}

\cvprfinalcopy 

\def\cvprPaperID{2480} 

\newcommand{\lspace}{\vspace{1mm}}

\begin{document}

\twocolumn
\setlength{\parindent}{1.0em}
\setlength{\parskip}{0ex plus 0.2ex minus 0.1ex}
\title{
Automating Carotid Intima-Media Thickness Video Interpretation with Convolutional Neural Networks\thanks{{\bf Shorter verson}: J. Y. Shin, N. Tajbakhsh, R. T. Hurst, C. B. Kendall, and J. Liang. Automating carotid intima-media thickness video interpretation with convolutional neural networks. IEEE Computer Society Conference on Computer Vision and Pattern Recognition (CVPR’16), Pages 2526-2535; {\bf Extended version}: N. Tajbakhsh, J. Y. Shin, R. T. Hurst, C. B. Kendall, and J. Liang. Automatic interpretation of carotid intima-media thickness videos using convolutional neural networks. Deep Learning for Medical Image Analysis. edited by Kevin Zhou, Hayit Greenspan and Dinggang Shen, Academic Press. 2017.}
}

\author{Jae Y. Shin$^{*}$, Nima Tajbakhsh$^{*}$,
        R. Todd Hurst, Christopher B. Kendall, and Jianming Liang
\thanks{J. Y. Shin, N. Tajbakhsh and J. Liang are with the Department
of Biomedical Informatics, Arizona State University, 13212 East Shea Boulevard, Scottsdale, AZ 85259, USA (e-mail: \{Sejong,Nima.Tajbakhsh, Jianming.Liang\}@asu.edu). Nima Tajbakhsh and Jae Y. Shin have contributed equally.}
\thanks{R.~T.~Hurst and C.~Kendall are with the Division of Cardiovascular Diseases of Mayo Clinic, 13400 E. Shea Blvd., Scottsdale, AZ 85259, USA (e-mail: \{Hurst.R, Kendall.Christopher\}@mayo.edu).}
}

\maketitle

\begin{abstract}
Cardiovascular disease (CVD) is the leading cause of mortality yet largely preventable, but the key to prevention is to identify at-risk individuals before adverse events. For predicting individual CVD risk, carotid intima-media thickness (CIMT), a noninvasive ultrasound method, has proven to be valuable, offering several advantages over CT coronary artery calcium score. However, each CIMT examination includes several ultrasound videos, and interpreting each of these CIMT videos involves three operations: (1) select three end-diastolic ultrasound frames (EUF) in the video, (2) localize a region of interest (ROI) in each selected frame, and (3) trace the lumen-intima interface and the media-adventitia interface in each ROI to measure CIMT. These operations are tedious, laborious, and time consuming, a serious limitation that hinders the widespread utilization of CIMT in clinical practice. To overcome this limitation, this paper presents a new system to automate CIMT video interpretation. Our extensive experiments demonstrate that the suggested system significantly outperforms the state-of-the-art methods. The superior performance is attributable to our unified framework based on convolutional neural networks (CNNs) coupled with our informative image representation and effective post-processing of the CNN outputs, which are uniquely designed for each of the above three operations.
\end{abstract}

\begin{figure}
\centering
\subfloat{\includegraphics[width=1.0\linewidth]{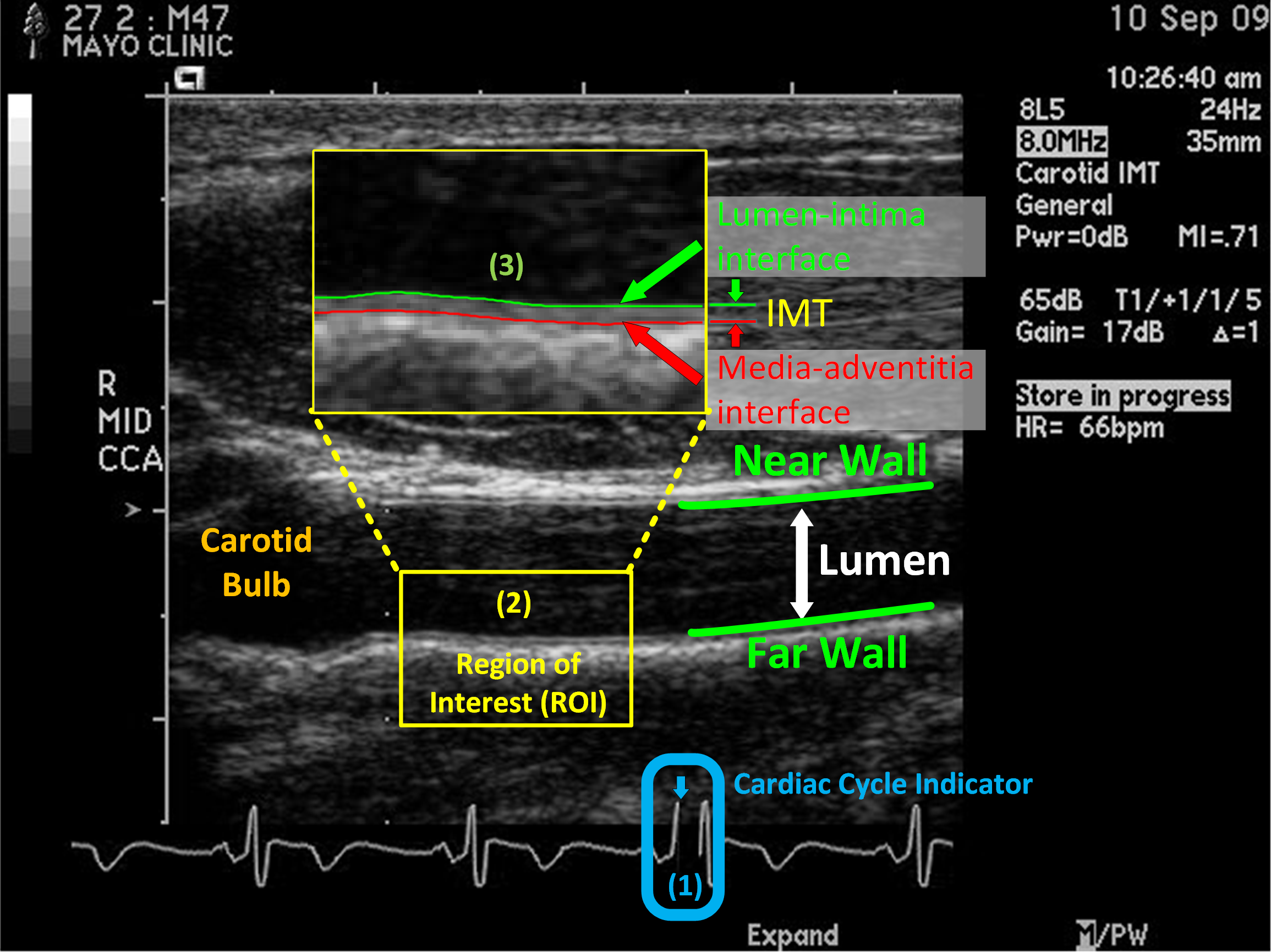}}

\caption{Longitudinal view of the carotid artery in an ultrasound B-scan image. CIMT is defined as the distance between the lumen-intima interface and the media-adventitia interface, measured approximately \SI{1}{\centi\metre} distal from the carotid bulb on the far wall of the common carotid artery at the end of the diastole; therefore, interpreting a CIMT video involves three operations: (1) select three end-diastolic ultrasound frames (EUFs) in each video (the cardiac cycle indicator, a black line, shows to where in the cardiac cycle the current frame corresponds); (2) localize a region of interest (ROI) approximately \SI{1}{\centi\metre} distal from the carotid bulb in the selected EUF; (3) measure the CIMT within the localized ROI. This paper aims to automate these three operations simultaneously through a unified framework based on convolutional neural networks.  }

\label{fig:intro}
\end{figure}

\section{Introduction}
\label{sec:RelatedWork}
Given the clinical significance of carotid intima-media thickness (CIMT) as an early and reliable indicator of cardiovascular risk, several methods have been developed for CIMT image interpretation.  The CIMT is defined as the distance between the lumen-intima and media-adventitia interfaces at the far wall of the carotid artery (\figurename~\ref{fig:intro}). Therefore, to measure CIMT, the lumen-intima and the media-adventitia interfaces must be identified. As a result, the earlier approaches are focused on analyzing the intensity profile and distribution, computing the gradient \cite{pignoli87, touboul92,faita08}, or combining various edge properties through dynamic programming \cite{liang00,cheng08,rossi10}. Recent approaches \cite{loizou07,delsanto07,petroudi12,xu12, ilea13, bastida13} are mostly based on active contours (aka, snakes) or their variations \cite{kass88}. Some of these approaches require user interaction, while other approaches aim for complete automation through integrating with various image processing algorithms, such as Hough transform \cite{molinari12} and dynamic programming \cite{rossi10}. Most recently, Menchón-Lara et al.\ employed a committee of standard multilayer perceptrons in \cite{menchon13} and a single standard multilayer perceptron with an auto-encoder in \cite{menchon15} for CIMT image interpretation, but both methods did not outperform  the snake-based methods from the same research group \cite{bastida13,bastida15}. For a more complete survey of methods for automatic CIMT measurements, please refer to the review studies conducted by Molinari et al.\cite{molinari10} and Loizou et al. \cite{loizou14}. 

\begin{figure*}
\centering
\subfloat{\includegraphics[width=.8\linewidth]{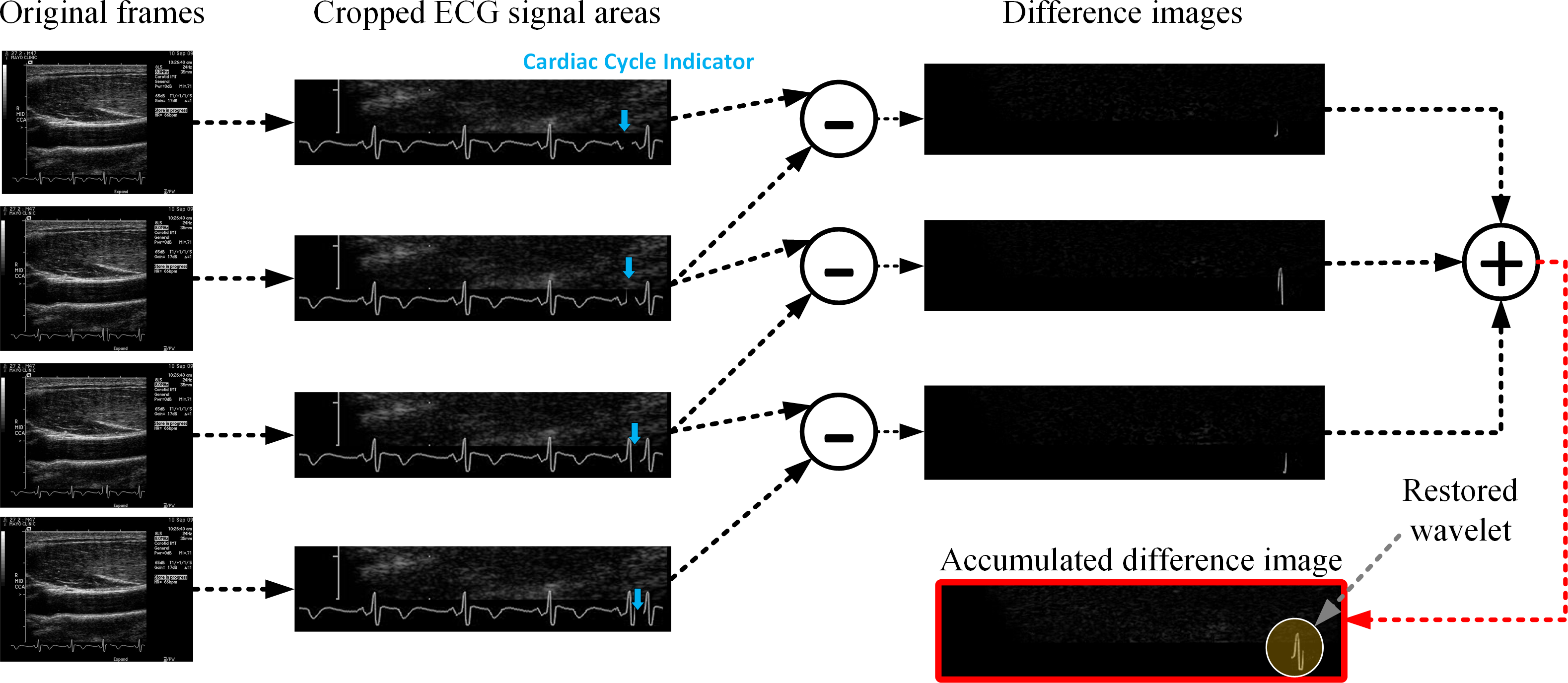}}\\
\caption{An accumulated difference image is generated by adding up three neighboring difference images.}
\label{fig:AccDiffIM}
\end{figure*}


However, nearly all the aforementioned methods are focused on only the third operation: CIMT measurement, ignoring the two preceding operations, i.e., frame selection and ROI localization. To our knowledge, the only system that simultaneously automates the three operations is {\iffalse our \else the \fi} work \cite{sharma14}, an extension of \cite{zhu11}, which automatically selects the EUF frame, localizes the ROI in each selected EUF frame, and provides the CIMT measurement in the selected ROI. However, as with other works, this method is based on hand-crafted algorithms, which often lack the desired robustness for routine clinical use, a weakness that we aim to overcome in this paper.  

A key contribution of this paper is a new system that accelerates CIMT video interpretation by automating all the three operations in a novel unified framework based on convolutional neural networks (CNNs). 
We will show that with proper pre-processing and post-processing, our proposed CNN-based approach can significantly outperform the existing methods in all aspects of CIMT image interpretation including frame selection, ROI localization, and CIMT measurements, making the following specific contributions:
\begin{itemize}
 \setlength\itemsep{-1mm}
\item	A unified framework based on CNNs that automates the entire CIMT interpretation process. This is in contrast to the prior works where only the very last step of the CIMT interpretation process was automated. The performance of the suggested system significantly outperforms the hand-crafted approach~\cite{sharma14}, which, to our knowledge, is the only system in the literature that aimed to automate all the above three tasks.
\item A novel frame selection method based on the ECG signals at the bottom of ultrasound frames. The suggested method utilizes effective pre-processing of patches and post processing of CNN outputs, enabling a significant increase in the performance of a baseline CNN.
\item A new  method that localizes the ROI for CIMT interpretation. The suggested method combines the discriminative power of a CNN with a contextual constrain to accurately localize the ROIs in the selected frames. We demonstrate that the suggested contextually-constrained CNN outperforms the performance of a baseline CNN.
\item A framework that combines CNNs with active contour models for accurate boundary segmentation. Specifically, given a localized ROI,  the CNN initializes two open snakes, which further deform to acquire the shapes of intima-media boundaries. We show that the segmentation accuracy of the suggested method is far higher than the state-of-the-art methods. 
\item Extensive evaluation of each stage of the suggested CIMT interpretation system. Specifically, we perform leave-one-patient-out cross-validation\footnote{We leave all the videos from one patient out for validation.} using only the training CIMT videos to tune the parameters of the suggested system, and then thoroughly evaluate the performance of our system using a large number of independent test CIMT videos.
\end{itemize}

\section{CIMT Protocol}


The CIMT exams utilized in this paper were performed with B-Mode ultrasound using an 8-14MHz linear array transducer utilizing fundamental frequency only (Acuson Sequoia\textsuperscript{TM}, Mountain View, CA, USA)~\cite{hurst10}. The carotid screening protocol begins with scanning bilateral carotid arteries in a transverse manner from the proximal aspect to the proximal internal and external carotid arteries. The probe is then turned to obtain the longitudinal view of the distal common carotid artery. The sonographer optimizes the 2D images of the lumen-intima and media-adventitia interfaces at the level of the common carotid artery by adjusting overall gain, time gain, compensation and focus position. Once the parameters are optimized, the sonographer captures two CIMT videos focused on the common carotid artery from two optimal angles of incidence. The same procedure is repeated for the other side of neck, resulting in a total of 4 CIMT videos for each subject.


\section{Method}
\label{method}




Our goal is to automate the three operations in CIMT video interpretation, i.e, given a CIMT video, our method will automatically identify three EUFs (Section \ref{sec:fs}), localize an ROI in each EUF (Section \ref{sec:roi}), and segment the lumen-intima and media-adventitia interfaces within each ROI (Section \ref{sec:imt}). 

\subsection{Frame Selection}
\label{sec:fs}



We select the EUFs based on the ECG signal embedded at the bottom part of a CIMT video. The cardiac cycle indicator is represented by a moving  black line in each frame.  Since the ECG signal is overlaid on the ultrasound image, there is quite bit of noise around the indicator.  The challenge is to reconstruct the original ECG signal from noisy frames and to detect the R peaks from the ECG signal, as the R-peaks correspond to the EUFs. To do so, we introduce accumulated difference images that carry sufficient information for CNN to learn and distinguish R-peaks from non-R-peaks.

\lspace
\noindent {\bf Training Phase:} Let $I^t$ denote an image subregion selected from the lower part of an ultrasound frame so that it contains the ECG signal. We first construct a set of difference images $d^{t}$ by subtracting every consecutive pairs of images, $d^t=|I^t- I^{t+1}|$, and then form accumulated difference images by adding up every three neighboring
difference images, $D^t= \sum_{i=0}^{2}d^{t-i}$. Accumulated difference image $D^t$  can capture the cardiac cycle indicator at frame $t$. \figurename~\ref{fig:AccDiffIM} illustrates how an accumulated difference image is generated. 

Next, we determine the location of the restored wavelet in each accumulated difference image.  For this purpose, we find the weighted centroid $c=[c_x,c_y]$ of each accumulated difference image $D^t$ as follows:

$$ c=\frac{1}{Z_t} \sum_{p\in D^t}{D^t(p_x,p_y)\times p} $$

\noindent where $p=[p_x,p_y]$ is a pixel in the accumulated difference image and $Z_t=\sum_{p\in D^t}{D^t(p_x,p_y)}$ is a normalization factor that ensures the weighted centroid stays within the image boundary. Once centroids are identified, we extract patches of size $32\times 32$ around the centroid locations. Specifically, we extract patches with up to 2 pixel translations from each centroid. However, we do not scale the patches in data augmentation, because doing so would inject label noise in the training set.  For instance, a small restored wavelet may take the appearance of an R-peak after expanding or an R-peak wavelet may look like a non-R-peak wavelet after shrinking. Nor do we perform rotation-based patch augmentation, because we do not expect the restored wavelets to appear with rotation in the test image patches. Once collected, patches are binarized using Otsu's method. In Section \ref{ex:fd}, we discuss the choice of binarization method through an extensive set of experiments.  Each binary patch is then labeled as positive if it corresponds to an EUF (i.e., an R-peak); otherwise negative. Basically, given a patch, we first determine the accumulated difference image from which the patch is extracted. We then trace back to the underlying difference images and check whether they are related to the EUF or not. Once the patches are labeled, we form a stratified set with 96,000 patches to train a 2-way CNN for frame selection. 


\begin{figure}
\centering
\subfloat{\includegraphics[width=1.0\linewidth]{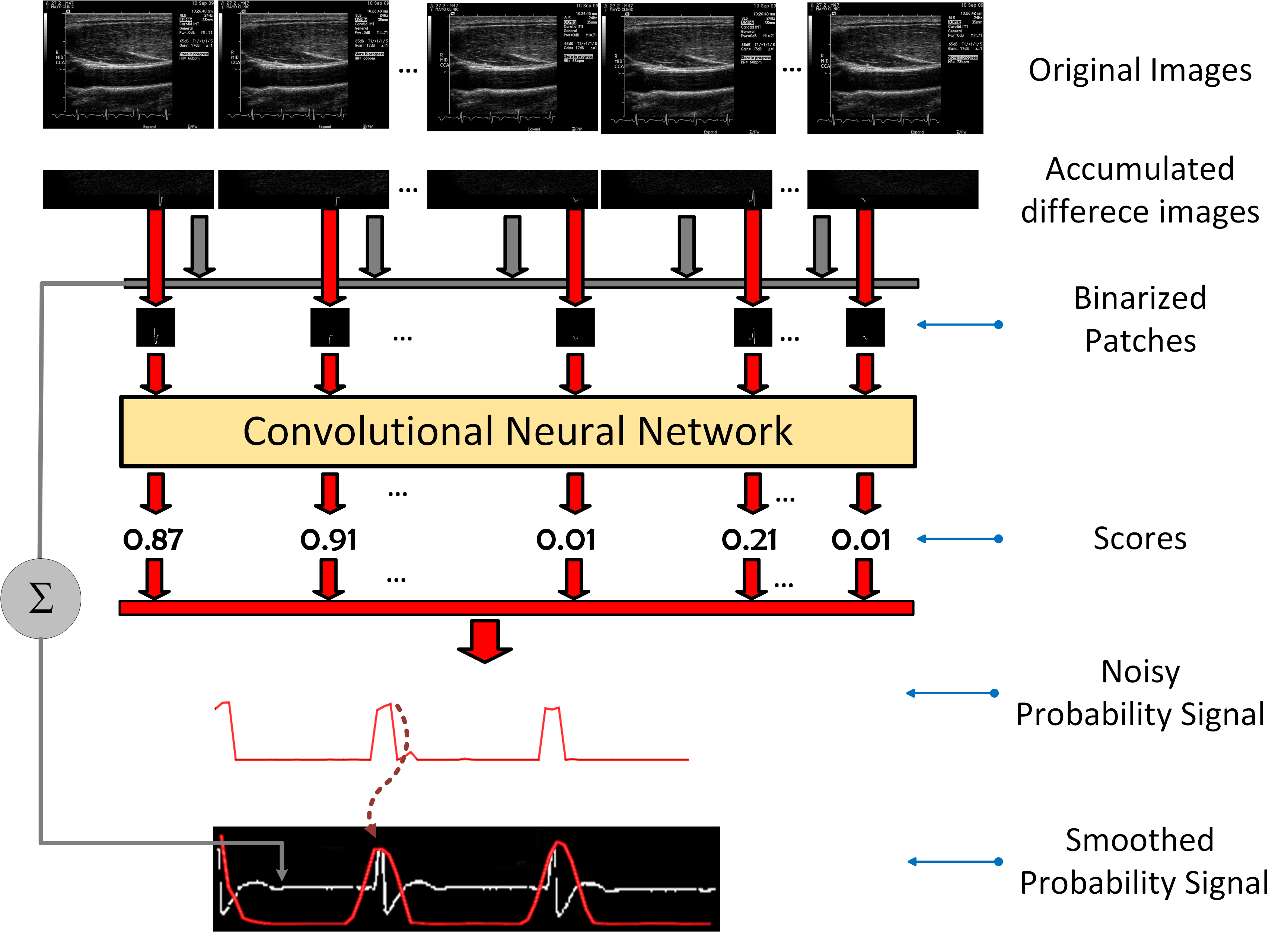}}
\caption{{The test stage of our automatic frame selection scheme. }}
\label{fig:fd_testStage}
\end{figure}

\lspace
\noindent {\bf Testing Phase: } \figurename~\ref{fig:fd_testStage} shows our frame selection system given a test video. We  first compute an accumulated difference image for each frame in the video. We then extract image patches from the weighted centroids of the accumulated difference images.  The probability of each frame being the EUF is measured as the average probabilities assigned by the CNN to the corresponding patches. By concatenating the resulting probabilities for all the frames in the video, we obtain a probability signal whose local maxima indicate the locations of the EUFs. However, the generated probability signals often exhibit abrupt changes, which can cause too many local maxima along the signal. We therefore first smooth the probability signal using a Gaussian function, and then find the EUFs by locating the local maxima of the smoothed signals. In \figurename~\ref{fig:fd_testStage}, for illustration purposes, we have also shown the reconstructed ECG signal, which is computed as the average of the accumulated difference images, $\frac{1}{N}\sum_{t=1}^N{D^t}$ with $N$ being the number of frames in the video. As seen, the probability of being the EUF reaches its maximum around the R peaks of the QRS complexes (as desired) and then smoothly decays as it distances from the R peaks. By mapping the locations of the local maxima to the frame numbers, we can identify the EUFs in the test video.

\subsection{ROI Localization}
\label{sec:roi}


Accurate localization of the ROI is challenging, because, as seen in \figurename~\ref{fig:intro}, there are no significant differences that can be observed in image appearance among the ROIs on the far wall of the carotid artery. To overcome this challenge, we utilize the location of the carotid bulb as a contextual constraint. We choose this constraint for two reasons: 1) the carotid bulb appears as a distinct dark area in the ultrasound frame and thus can be uniquely identified; 2) according to the consensus statement of American society of Electrocardiography for cardiovascular risk assessment, the ROI should be placed approximately \SI{1}{\centi\metre} from the carotid bulb on the far wall of the common carotid artery. While the former motivates the use of the carotid bulb location as a constraint from a technical point of view, the latter justifies this constraint from a clinical standpoint. 


\lspace
\noindent {\bf Training Phase:} We incorporate this constraint in the suggested system by training  a 3-way CNN that simultaneously localizes both ROI and carotid bulb, and then refines the estimated location of the ROI given the location of the carotid bulb. {\figurename~\ref{fig:cb_roi} in the supplementary material} illustrates how the image patches are extracted from a training frame. We perform data augmentation by extracting the training patches within a circle around the locations of the carotid bulbs and the ROIs. The negative patches are extracted from a grid of points sufficiently far from the locations of the carotid bulbs and the ROIs. Note that the above translation-based data augmentation is sufficient for this application, because our database provides a relatively large number of training EUFs, from which a \iffalse massive \else large \fi set of training patches can be collected. Once the patches are collected, we form a stratified training set with approximately 410,000 patches to train a 3-way CNN for constrained ROI localization.


\begin{figure*}
\centering
\subfloat{\includegraphics[width=0.9\linewidth]{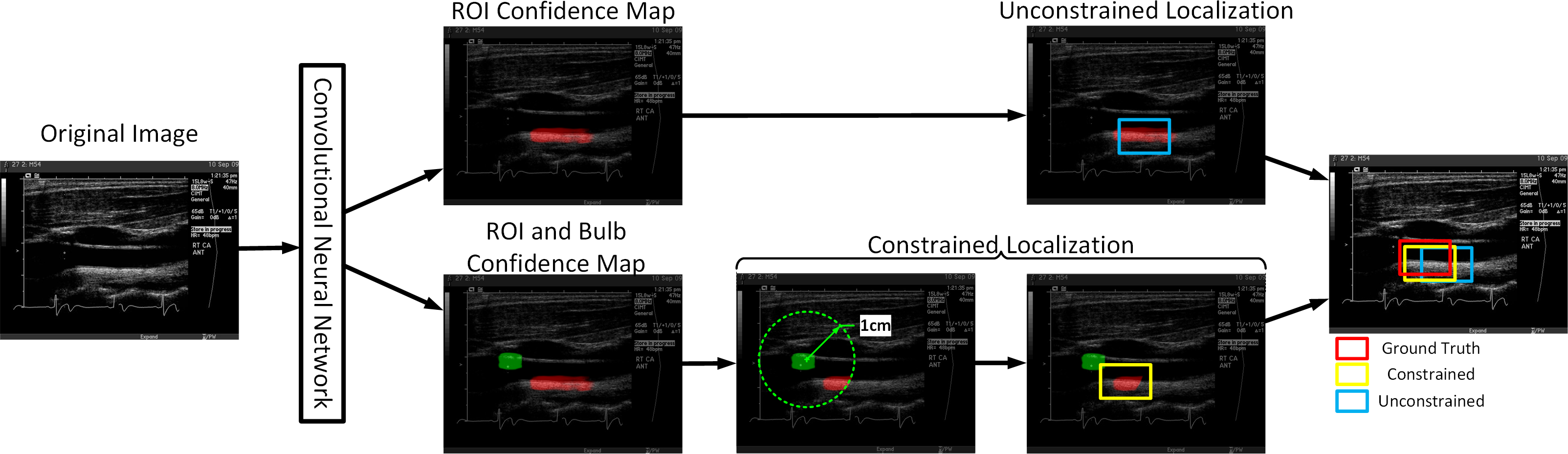}}
\caption{The test stage of our ROI localization method. In the unconstrained scenario, we only use  the ROI confidence map, which results in relatively large localization error. In the constrained mode, given the estimated location of the carotid bulb, we localize the ROI more accurately.}
\label{fig:roiLocalization}
\end{figure*}

\lspace
\noindent {\bf Testing Phase:} Referring to \figurename~\ref{fig:roiLocalization}, during the test stage,  the trained CNN is applied to all the pixels in the EUF, generating two confidence maps with the same size as the EUF. The first confidence map shows the probability of a pixel being the carotid bulb and the second confidence map shows the probability of a pixel being the ROI. One way to localize the ROI is to find the center of the largest connected component within the ROI confidence map without considering the detected location of the carotid bulb. However, this naive approach may  fail to accurately localize the ROI. For instance, a long-tale connected component along the far wall of the carotid artery may cause substantial ROI localization error. To compound the problem, the largest  connected component of the ROI confidence map may appear far from the actual location of the ROI, resulting in a complete detection failure. To overcome these limitations, we constraint the ROI location $l_{roi}$ by the location of the carotid bulb $l_{cb}$. For this purpose, we first determine the location of the carotid bulb as the centroid of the largest  connected component within the first confidence map, and then localize the ROI using the following formula 


\begin{equation} 
l_{roi} = \frac{\sum_{p\in C^*}{M(p)\cdot p \cdot I(p)}}{\sum_{p\in C^*}{M(p) \cdot I(p)}}
\label{eq:roi}
\end{equation}

\noindent where $l_{roi}$ denotes the ROI location, $l_{cb}$ denotes the center of the carotid bulb, $M$ denotes the confidence map of being the ROI, $C^*$ is the largest  connected component in $M$ that is the nearest to the carotid bulb, and I(p) is an indicator function for pixel $p=[p_x,p_y]$ that is defined as


\begin{empheq}[left={I(p)=\empheqlbrace}]{align}
    1, &\quad \text{if } \|p-l_{cb}\|<1\, \text{cm} \label{eq:roi_indicator_1}\\
    0, &\quad \text{otherwise} \label{eq:roi_indicator_0}
\end{empheq}


The indicator function I(p) is binary function that simply includes pixel when the value is 1 as in  Eq.~\ref{eq:roi_indicator_1}, otherwise excludes pixel when the value is 0 as in Eq.~\ref{eq:roi_indicator_0}.

\subsection{Intima-Media Thickness Measurement}
\label{sec:imt}


Measuring intima-media thickness require a continuous and one-pixel precise boundary for lumen-intima and media-adventitia.  Lumen-intima is relatively easier to detect because of strong gradient change at the border, however, detecting media-adventitia interface is quite challenging due to its subtle image gradients and noise around its border.  We approach this problem as a 3-way classification task: 1) lumen-intima interface, 2) media-adventitia interface, and 3) background.


\lspace
\noindent {\bf Training Phase:}
To train 3-way CNN, we collected sparse background patches and then pixel-by-pixel image patches around lumen-intima interface and media-adventitia interface with additional patches $\pm$3 pixels from the ground truth.  Using $\pm$3 pixels for additional patches around intima-media boundary was necessary to balance number of patches with background patches and produced better results. {\figurename~\ref{fig:cimt_patch} in the supplementary material} illustrates how the training patches are collected from an ROI.

\begin{figure*}
\centering
\subfloat{\includegraphics[width=0.8\linewidth]{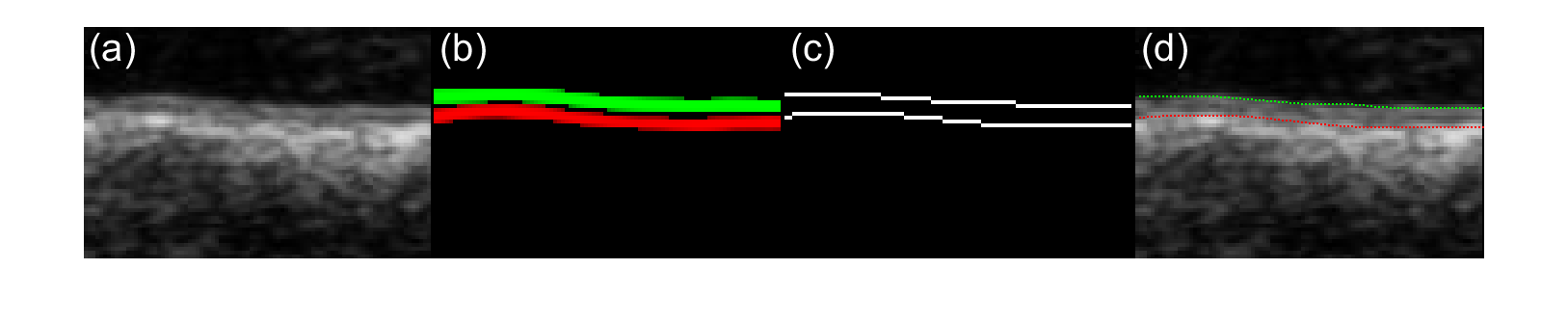}}
\caption{The test stage of lumen-intima and media-adventitia interface detection. (a) a test ROI. (b) The trained CNN generates a confidence map where the green and red colors indicate the likelihood of lumen-intima interface and media-adventitia interface, respectively. (c) The thick probability band around each interface is thinned by selecting the largest probability for each interface in each column. (d) The step-like boundaries are refined through two open snakes.}
\label{fig:cimt_workflow}
\end{figure*}

\lspace
\noindent {\bf Testing Phase:}
\figurename~\ref{fig:cimt_workflow} illustrates the testing process.  The 3-way trained CNN is applied in a sliding-window fashion for a given test ROI and generates two confidence maps (\figurename~\ref{fig:cimt_workflow}(b)) with the same size as the ROI.  Since confidence map is thicker than a pixel, we choose the maximum response column-by-column and generate a new binary image as shown in \figurename~\ref{fig:cimt_workflow}(c).  Finally, we use two active contour models (a.k.a, snakes) \cite{liang06} for segmenting lumen-intima and media-adventitia interfaces.  \figurename~\ref{fig:cimt_workflow}(d) shows two final converged snakes and we take measurements as the average vertical distance between the two snakes.

\section{Experiments}
\label{sec:exp}


We use a database of 92 CIMT videos captured from 23 subjects with 2 CIMT videos from the left and 2 CIMT videos from the right carotid artery of each subject. The ground truth for each video contains the EUF number, the locations of ROI, and the segmentation of lumen-intima and media-advantitia interfaces. For consistency, we use the same training set and the same test set (no overlap with training) for all three tasks. Our training set contains 48 CIMT videos of 12 subjects with a total of 4,456 frames and our test set contains 44 CIMT videos of 11 subjects with a total of 3,565 frames. For each task, we perform leave-one-patient-out cross-validation based on the {\em training subjects} to tune the parameters, and then evaluate the performance of the tuned system using the test subjects. 


\lspace
\noindent {\bf Architecture:} \nocite{le90} As shown in Table~\ref{fig:netArch}, we employ a CNN architecture with 2 convolutional layers, 2 subsampling layers, and 2 fully connected layers (see Section~\ref{sec:discussion} for our justifications). We also append a softmax layer to the last fully connected layer so as to generate probabilistic confidence values for each class. Our CNN architecture has input patches of size 32x32, and we resize the collected patches to 32x32 prior to the training process. For the CNNs used in our experiments, we employ a learning rate of $\alpha=0.001$, a momentum of $\mu=0.9$, and a constant scheduling rate of $\gamma=0.95$.

%
\begin{table*}
\caption{The CNN architecture used in our experiments. Note that $C$ is the number of classes, which is 2 for frame selection and 3 for both ROI localization and intima-media thickness measurements.}
\begin{center}
\begin{tabular}{*{7}{|c|c|c|c|c|c|c}}

\cline{1-7}
layer&type&input&kernel&stride&pad&output\\
\hline
0& input  & 32x32 & N/A & N/A &N/A & 32x32 \\[.1cm]
1& convolution  & 32x32 & 5x5 & 1 & 0 & 64x28x28\\[.1cm]
2& max pooling  & 64x28x28 & 3x3 & 2 & 0 & 64x14x14\\[.1cm]
1& convolution  & 64x14x14 & 5x5 & 1 & 0 & 64x10x10\\[.1cm]
2& max pooling  & 64x10x10 & 3x3 & 2 & 0 & 64x5x5\\[.1cm]
2& fully connected  & 64x5x5 & 5x5 & 1 & 0 & 250x1\\[.1cm]
2& fully connected  & 250x1 & 1x1 & 1 & 0 & Cx1\\[.1cm]
\hline
\end{tabular}
\end{center}
\label{fig:netArch}
\end{table*}

\lspace
\noindent {\bf Pre- and post-processing for frame selection:}
\label{ex:fd}
We have experimentally found out that binarized image patches improve the quality of convergence and accuracy of frame selection. Furthermore, we have observed that the standard deviation of the Gaussian function used for smoothing the probability signals, can also substantially influence frame selection accuracy. Therefore, we have conducted leave-one-patient-out cross-validation based on the training subjects to find the best binarization method and the optimal standard deviation of the Gaussian function. For binarization, we have considered a fixed set of thresholds and  adaptive thresholding using Otsu's method. For smoothing, we have considered a Gaussian function with different standard deviation ($\sigma_g$) as well as the scenario where no smoothing is applied. For each configuration of parameters, we have done a free-response ROC (FROC) analysis. We consider a selected frame a true  positive, if it is found within one frame from the expert-annotated EUF; otherwise, a false positive.

\begin{figure}
\centering
\subfloat{\includegraphics[width=.9\linewidth]{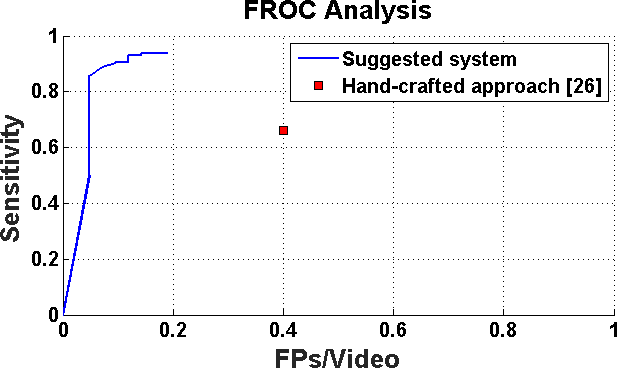}}
\caption{FROC curve of our frame selection system for the test subjects using the tuned parameters. For comparison, we have also shown the operating point of {\iffalse our \else the\fi} prior hand-crafted approach~\cite{sharma14}, which is significantly outperformed by the suggested system. }
\label{fig:frDetResults_te}
\end{figure}

{Our leave-one-patient-out cross-validation study, summarized in \figurename~\ref{fig:frDetResults_tr} in the supplementary material, indicates that the use of a  Gaussian function with $\sigma_g=1.5$ for smoothing the probability signals and  adaptive thresholding using Otsu's method achieve the highest performance.} \figurename~\ref{fig:frDetResults_te} shows the FROC curve of our system for the test subjects using the above parameters. For comparison, we have also shown the operating point of {\iffalse our \else the\fi} hand-crafted approach~\cite{sharma14}, which is significantly outperformed by the suggested system.

\begin{figure}
\centering
\subfloat{\includegraphics[width=.8\linewidth]{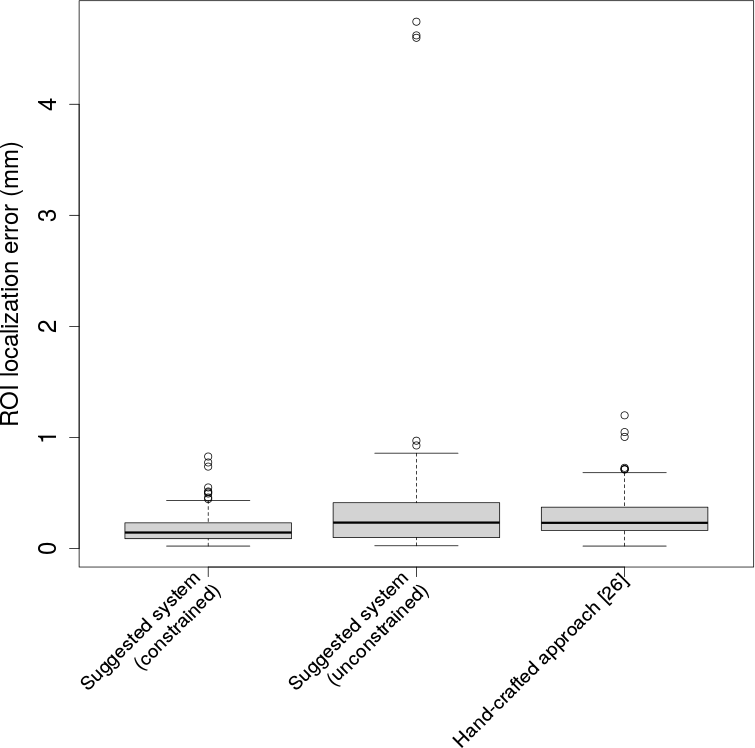}}
\caption{ROI localization error for the test subjects. Our method in the constrained mode outperforms both the unconstrained counterpart and {\iffalse our \else the\fi} prior hand-crafted approach~\cite{sharma14}.}
\label{fig:boxplot_roi_test}
\end{figure}

\lspace
\noindent {\bf Constrained ROI Localization:}
\label{ex:roi}
We conduct a leave-one-patient-out cross-validation study based on the training subjects to find the optimal size of the training patches. Our cross-validation analysis, summarized in \figurename~\ref{fig:boxplot_roi_cv} in the supplementary material, indicates that the use of $1.8\times1.8$ \SI{}{\centi\metre} patches achieves the most stable performance, yielding low   ROI localization error with only a few outliers.  \figurename~\ref{fig:boxplot_roi_test} shows the ROI localization error of our system for the test subjects using the optimal size of training patches. To demonstrate the effectiveness of our constrained ROI localization method, we have also included the performance of the unconstrained counterpart. In the constrained mode, we use Eq.~\ref{eq:roi} for ROI localization whereas in the unconstrained mode we localize the ROI as the center of the largest  connected component in the corresponding confidence map without considering the location of the carotid bulb. Our method achieves an average localization error of \SI{0.19}{\milli\metre} and \SI{0.35}{\milli\metre} in the constrained and unconstrained modes, respectively. The decrease in localization error is statistically significant ($p<0.01$). Also as seen in \figurename~\ref{fig:boxplot_roi_test}, our method in the unconstrained mode has resulted in 3 complete localization failures (outliers), which have been corrected in the constrained mode. Furthermore, compared with {\iffalse our \else the\fi} hand-crafted approach~\cite{sharma14}, our system in the constrained mode shows a decrease of \SI{0.1}{\milli\metre} in ROI localization error, which is statistically significant ($p<.00001$).

\label{ex:imt}

\lspace
\noindent {\bf Intima-Media Thickness Measurement:}
We determined the optimal image patch size by leave-one-patient-out cross-validation using various image patch sizes and found that $360\times360$ \SI{}{\micro\metre} achieved slightly lower localization error and fewer outliers (see Figs.~\ref{fig:boxplot_roi_cv}-\ref{fig:boxplot_IMT_cv} in the supplementary material). \figurename~\ref{fig:boxplot_te_imt} shows the interface localization error of our system on the test subjects, where we break down the overall localization error for lumen-intima and that of the media-adventitia interface as well as {\iffalse our \else the\fi} hand-crafted approach~\cite{sharma14} for each interface.
We further analyzed agreement between our system and the expert with the Bland-Altman plot (see \figurename~\ref{fig:blandAltman_te_imt_jimmy} in the supplementary material).
\begin{figure}
\centering
\subfloat{\includegraphics[width=.8\linewidth]{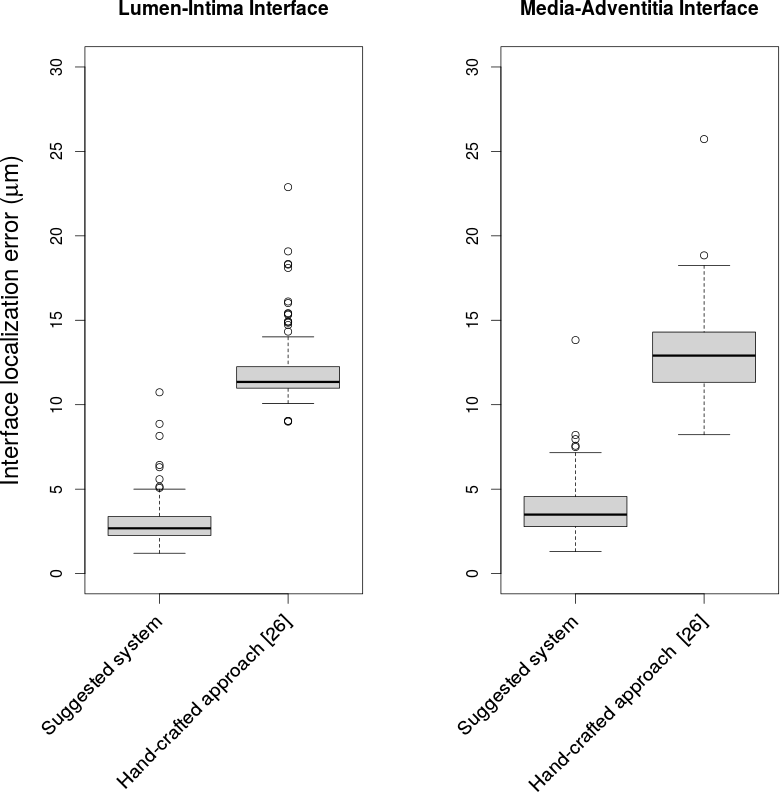}}
\caption{Localization error of the lumen-intima and media-adventitia interfaces for the suggested system and {\iffalse our \else the\fi} prior hand-crafted approach~\cite{sharma14}. The results are obtained for the test subjects.}
\label{fig:boxplot_te_imt}
\end{figure}

\section{Discussions}
\label{sec:discussion}



 In Section~\ref{ex:fd}, we investigated how the choice of patch binarization and degree of Gaussian smoothing affect the accuracy of frame selection. Here, we would like to discuss our findings and provide insights about our choices. We choose to binarize the patches, because it reduces appearance variability and suppress the low-magnitude noise content in the patches. Without patch binariztion, one can expect a large amount of variability in the appearance of wavelets can deteriorate the performance of the subsequent CNN (see \figurename~\ref{fig:frDetResults_tr} in the supplementary material). The choice of binarization threshold is another important factor. The use of a high threshold results in the partial appearance of the wavelets in the resulting binary patches, reducing the discriminatory appearance features of the patches.  A low threshold, on the other hand, can intensify noise content in the images, which decreases the quality of training samples and consequently a drop in classification performance. According to our analyses, it is difficult to find a fixed threshold that can both suppress the noise content and keep the shapes of the restored wavelets intact in all the collected patches. Otsu's method seems to overcome this limitation by adaptively selecting a binarization threshold according to the intensity distribution of each individual patch. For patches with intensity values between 0 and 1, the adaptive thresholds have a mean of 0.15 and standard deviation of 0.05. The wide range of adaptive thresholds explains why a constant threshold may not perform as desirably.

Gaussian smoothing of the probability signals is also essential for accurate frame selection. This is because the probability signals prior to smoothing exhibit high frequency fluctuations, which may complicate the localization of the local maxima in the signals. The first cause of such high frequency changes  is patch misplacement in the accumulated difference images.  Recall that we extract the patches around the weighted centroids of the accumulated difference images. However, a large amount of noise content in the difference images may cause the weighted centroid to deviate from the center of the restored wavelet. In this case, the extracted patch may partially or completely miss the restored wavelet. This can manifest itself as a sudden change in the CNN output and as a result in the corresponding probability signal. The second cause of high frequency changes is the inherited high variance of CNNs. Use of ensemble of CNNs and data augmentation can alleviate this problem at a significant computation cost. Alternatively, we choose to mitigate these undesirable fluctuations using Gaussian smoothing \iffalse, which allows for both time and \else for \fi computational efficiency.

As described in Section~\ref{sec:roi}, we constrain our ROI localization method by the location of the carotid bulb. This is because the bulb area appears as a relatively distinct  dark area in the ultrasound frame. The distinct appearance of the carotid bulb is also confirmed by our experiments, where we obtain the average bulb localization error of \SI{0.26}{\milli\metre} for the test subjects with only one failure case, which is more favorable than the average unconstrained ROI localization error of \SI{0.38}{\milli\metre} with 3 failure cases. Therefore, the localization of the bulb area can be done more reliably than the localization of the ROI, which motivates  the use of the bulb location  as a guide for more accurate ROI localization.  We integrate this constraint into our localization system through a post-processing mechanism (see Eq.~\ref{eq:roi}). Alternatively, we could train a regression CNN where each pixel in the image directly  votes for the location of the ROI. However, this approach may be hindered by lack of stable anatomical structures in noisy ultrasound images. We will explore a regression CNN for ROI localization as future work. 





In Section \ref{ex:imt}, we showed a high level of agreement between our system and the expert for the assessment of intima-media thickness. The suggested system achieves a mean absolute error of \SI{2.8}{\micro\metre} with a standard deviation of \SI{2.1}{\micro\metre} for intimia-media thickness measurements. However, this level of measurement error cannot hurt the interpretation of the vascular age, because there exists a minimum difference of \SI{400}{\micro\metre} between the average intima-media thickness of healthy and high-risk population (\SI{600}{\micro\metre} for healthy and $\geq1000$ \SI{}{\micro\metre}  for high-risk population) \cite{Jacoby04}. To further put the performance of our system into perspective, in Table~\ref{PreviousWorks}, we have compared the accuracy of intima-media thickness measurements produced by our system with those of the other automatic methods recently suggested in the literature. As seen, our method yields a lower level of mean absolute error and smaller standard deviation.

We used a LeNet-like CNN architecture in our study, but it does not limit the suggested framework to this
architecture. In fact, we have experimented with deeper CNN architectures such as AlexNet \cite{krizhevsky12} in both training and fine-tuning modes; however, we did not observe any significant performance gain. This was probably because the higher level semantic features detected by the deeper networks are not very relevant to the tasks in our CIMT applications.  Meanwhile, the concomitant computational cost of deep architectures may hinder the applicability of our system, because it lowers the speed---a key usability factor of our system. We also do not envision that a shallower architecture can offer the performance required for clinical practice. This is because a network shallower than the LeNet has only one convolutional layer and thus limited to learning primitive edge like features. Detecting the carotid bulb and the ROI, and segmenting intima-media boundaries are relatively challenging tasks, requiring more than primitive edge-like features. Similarly, for frame selection, classifying the restored wavelets into R-peak and non-R-peak categories is similar to digit recognition, for which LeNet  is a common choice of architecture. Therefore, LeNet-like CNN architecture seems to represent an optimal balance between efficiency and accuracy for CIMT video analysis.
 
\nocite{long14}

We should note that throughout this paper, all performance evaluations were performed without involving any user interactions. However, our goal is not to exclude the user (sonographer) from the loop rather to relieve him from the three tedious, laborious, and time consuming operations by automating them while still offering the user a highly, user-friendly interface to bring his indispensable expertise onto CIMT interpretation through refining the automatic results easily at the end of each of the automated operations. For instance, our system is expected to automatically locate a EUF within one frame, which is clinically acceptable,  but in case the automatic selected EUF is not the exact one as desired, the user can simply press an arrow key to move one frame forward or backward. From our experience, the automatically localized ROI is acceptable even if there is a small distance from the ground truth location, but the user still can easily drag the ROI and move it around as desired. Finally, in refining the automatically identified lumen-intima and media-adventitia interfaces, the original snake formulation comes with spring forces for user interaction \cite{kass88}, but given the small distance between the lumen-intima and media-adventitia interfaces, we have found that ``movable'' hard constraints as proposed in \cite{liang06} are far more effective than the spring forces in measuring CIMT.

\begin{table}
\centering
{\small 
\caption{ CIMT error for our system and the other state-of-the-art methods.}
\begin{tabular}{*{5}{c}}
\hline
\bf {Author}& &\bf{ Year} & & \bf{Thickness error (\SI{}{\micro\metre})} \\ 
\hline
Current work & & --- & & 2.8 $\pm$ 2.1\\ 
Bastida-Jumillacite~\cite{bastida15} & & 2015& &13.8$\pm$31.9  \\ 
Ilea~\cite{ilea13} & & 2013 & & 80 $\pm$ 40\\ 
Loizou~\cite{loizou13} & & 2013 & & 30 $\pm$ 30\\ 
Molinari~\cite{molinari11} & & 2011 & & 43 $\pm$ 93\\ 
\hline 

\end{tabular}

\label{PreviousWorks}
}
\end{table}

\section{Conclusion}

In this paper, we presented a unified framework to fully automate and accelerate CIMT video interpretation. Specifically, we suggested a computer-aided CIMT measurement system with three components: (1) automatic frame selection in CIMT videos,  (2) automatic ROI localization within the selected frames, (3) automatic intima-media boundary segmentation within the localized ROIs. We based each of the above components on a CNN with a LeNet-like architecture and then boosted the performance of  the employed CNNs with effective pre- and post-processing techniques. For frame selection, we demonstrated that how patch binarization as a pre-processing step and smoothing the probability signals as a post-processing step improve the results  generated by the CNN. For ROI localization, we experimentally proved that the location of the carotid bulb, as a constraint in a post-processing setting, significantly improves ROI localization accuracy. For intima-media boundary segmentation, we employed open snakes as a post processing step to further improve the segmentation accuracy. We compared the results produced by the suggested system with those of {\iffalse our \else the\fi} major prior works, demonstrating more accurate frame selection, ROI localization, and CIMT measurements. This superior performance is attributed to the effective use of CNNs coupled with  pre- and post- processing steps, uniquely designed for the three CIMT tasks.


{\small
\bibliographystyle{ieee}
\bibliography{RefJPaper}
}




\section*{Supplementary material}

 \subsection*{Convolutional Neural Networks}
 \label{sec:CNN}
As with multi-layer prceptrons, convolutional neural networks are trained using the back-propagation algorithm. If $D$ denotes a set of training images, $W$ denotes a matrix containing the weights of the convolutional layers, and $f_W(D^{(i)})$ denotes the loss for the $i^{th}$ training  image, the loss over the entire training set is then computed as

\begin{equation}
\mathcal{L}(W) = \frac{1}{|D|}\sum_{i}^{|D|}{f_W(X^{(i)})}
  \end{equation}
  
Gradient descent is commonly used for minimizing the above loss function with respect to the unknown weights $W$. However, the modern massively parallelized  implementations of CNNs  are limited by the amount of memory on GPUs; therefore, one cannot evaluate the loss function based on the entire training set $D$ at once. Instead, the loss function is approximated with the loss over the mini-batches of training images of size $N<<|D|$. A common choice of the mini-batch size is 128, which is a reasonable trade-off between loss noise suppression and memory management. Given the size of mini-batches, one can approximate the loss function as $\mathcal{L}(W) \approx \frac{1}{N}\sum_{i=1}^{N}{f_W(X^{(i)})}$, and iteratively update the weights of the network with the following equations:

$$\gamma_t=\gamma^{\floor*{\frac{tN}{|D|}}}$$
$$V_{t+1} = \mu V_t - \gamma_t\alpha\Delta L(W_t)$$
\begin{equation}
W_{t+1} = W_t + V_{t+1}
\label{eq:weightUpdate}
\end{equation}

\noindent where $\alpha$ is the learning rate, $\mu$ is the momentum that indicates the contribution of the previous weight update in the current iteration, and $\gamma$ is the scheduling rate that decreases learning rate $\alpha$ at the end of each epoch. 

\newpage

 \subsection*{Figures}
\begin{figure}[h]
\centering
\subfloat{\includegraphics[width=1.0\linewidth]{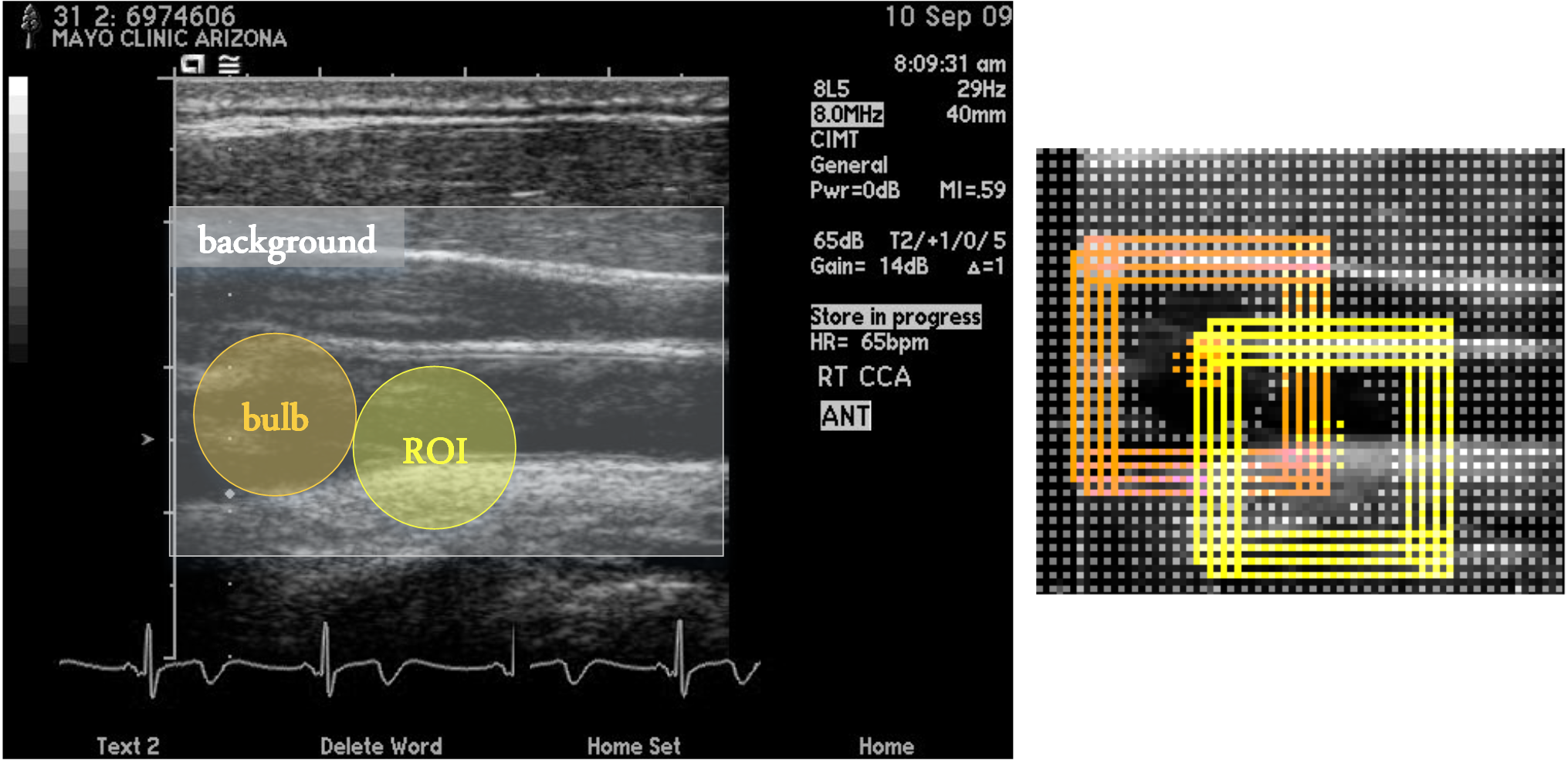}}
\caption{For constrained ROI localization, we use a 3-way CNN whose training image patches are extracted from a grid of points on the background and around the ROI and the carotid bulb locations.} 
\label{fig:cb_roi}
\end{figure}

\begin{figure}[h]
\centering
\subfloat{\includegraphics[width=1.0\linewidth]{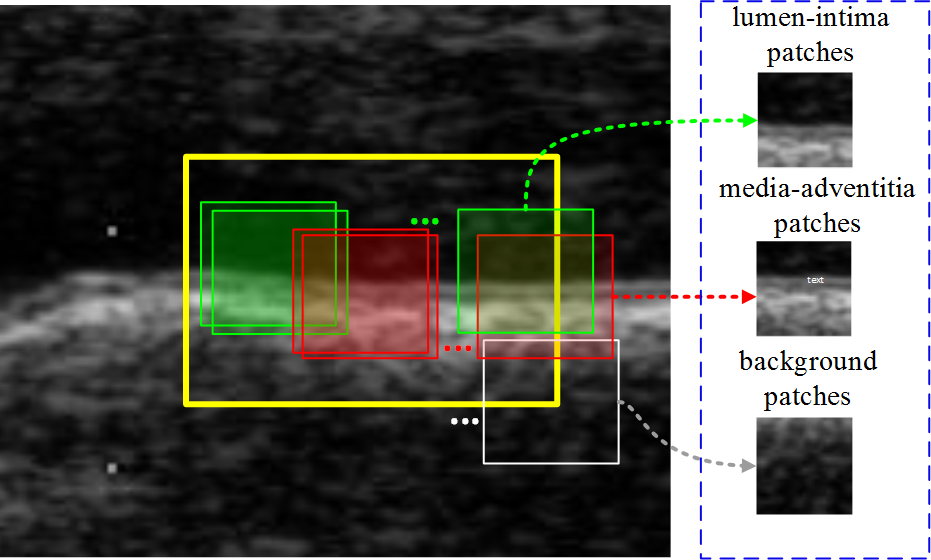}}
\caption{For lumen-intima and media-adventitia interface segmentation,  we use a 3-way CNN whose training image patches are extracted from the background and around the lumen-intima and media-adventitia interfaces.}
\label{fig:cimt_patch}
\end{figure}

\clearpage

\begin{figure*}[h]
\centering
\subfloat{\includegraphics[width=.32\linewidth]{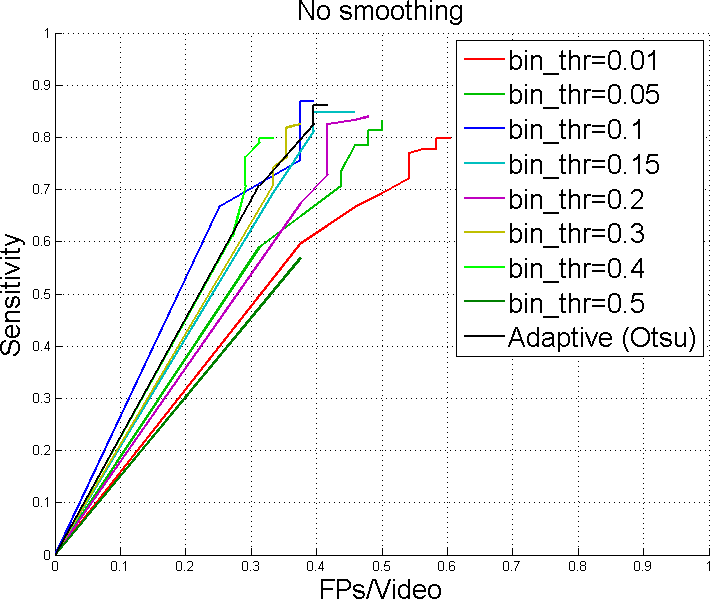}}\hspace{1pt}
\subfloat{\includegraphics[width=.32\linewidth]{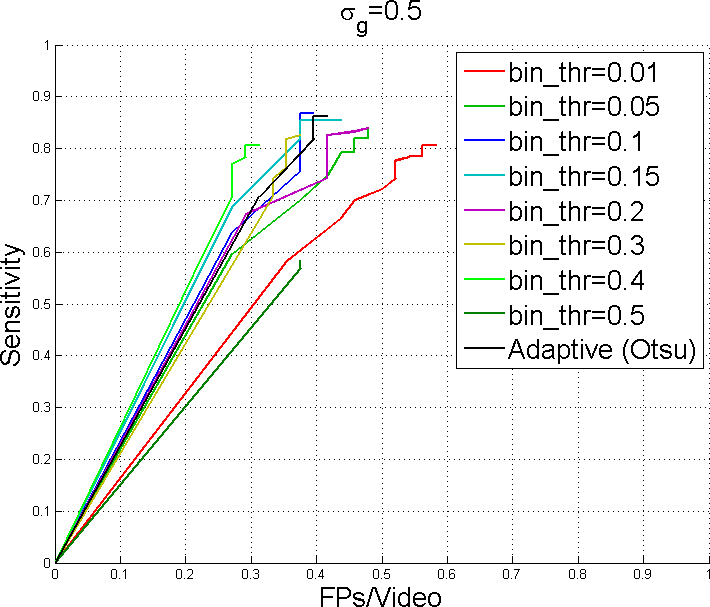}}\hspace{1pt}
\subfloat{\includegraphics[width=.32\linewidth]{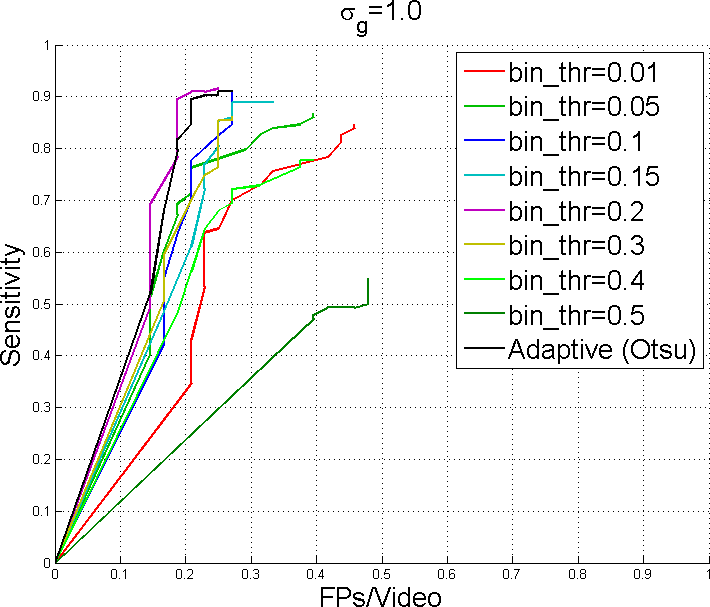}}\\
\subfloat{\includegraphics[width=.32\linewidth]{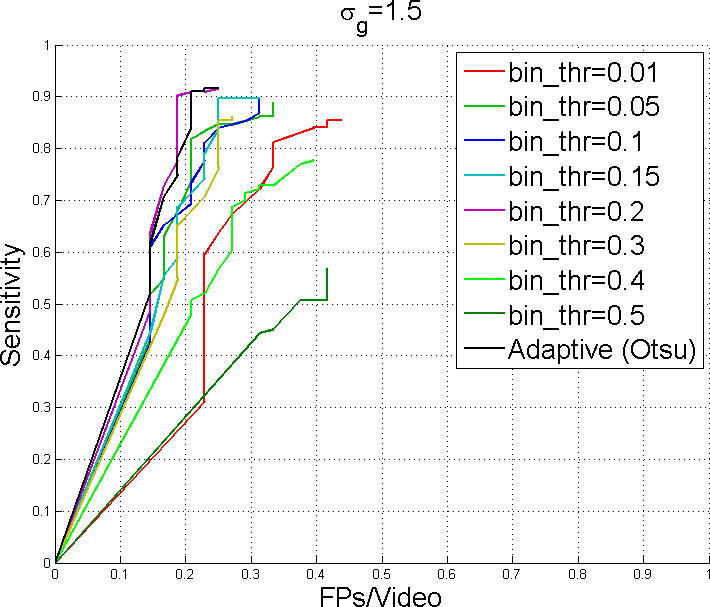}}\hspace{1pt}
\subfloat{\includegraphics[width=.32\linewidth]{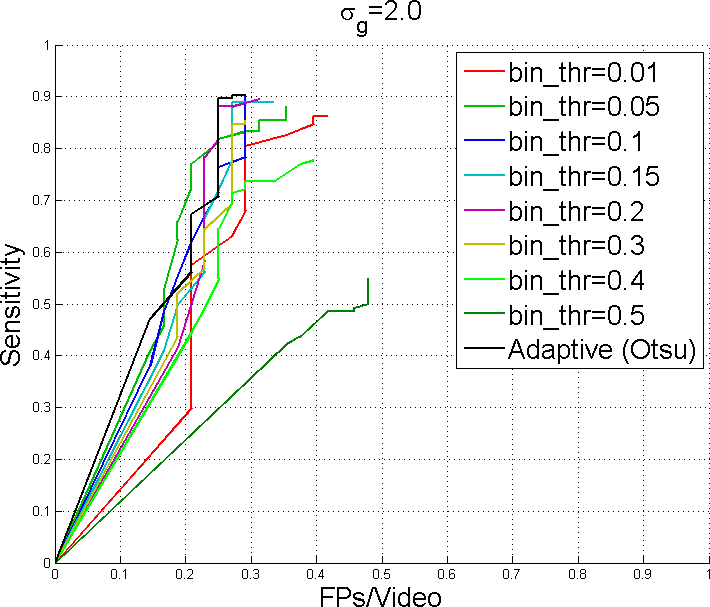}}\hspace{1pt}
\subfloat{\includegraphics[width=.32\linewidth]{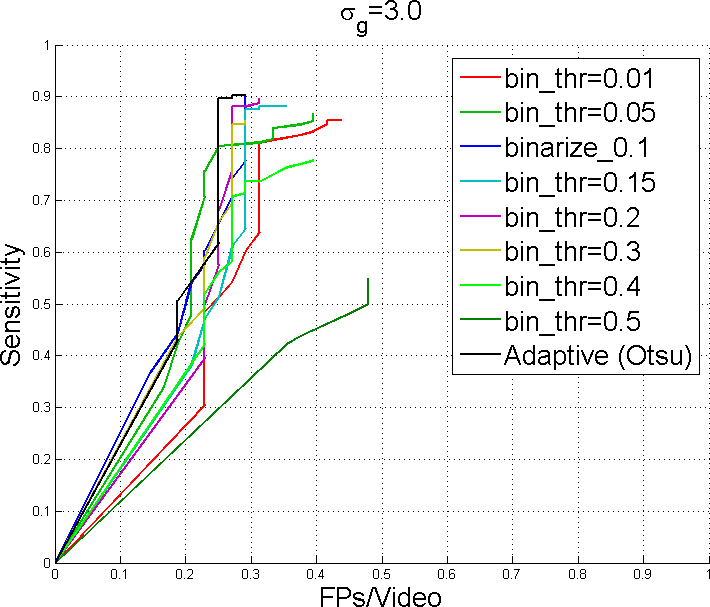}}

\caption{FROC curves of our system for automatic frame selection. Each plot shows FROC curves for different binarization thresholds and different levels of Gaussian smoothing. The results are obtained using leave-one-patient-out cross-validation based on the training subjects. As seen, no smoothing or a small degree of Gaussian smoothing leads to relatively low frame selection accuracy. This is because a trivial level of smoothing may not properly handle the fluctuations in the probability signals, causing a large number of false positives around an EUF. On the other hand, a large degree of smoothing may decrease the sensitivity of frame selection as the locations of the local maxima may be found more than one frame away from the expert-annotated EUFs. We therefore use a Gaussian function with $\sigma_g=1.5$ for smoothing the probability signals. Our results also indicate that the adaptive thresholding method and a fixed threshold of 0.2 achieve the highest frame selection accuracy. However, we choose to use adaptive thresholding because it decreases the parameters of our system by one and that it performs more consistently at different levels of Gaussian smoothing.}
\label{fig:frDetResults_tr}
\end{figure*}

\clearpage

\begin{figure}[h]
\centering
\subfloat{\includegraphics[width=1.0\linewidth]{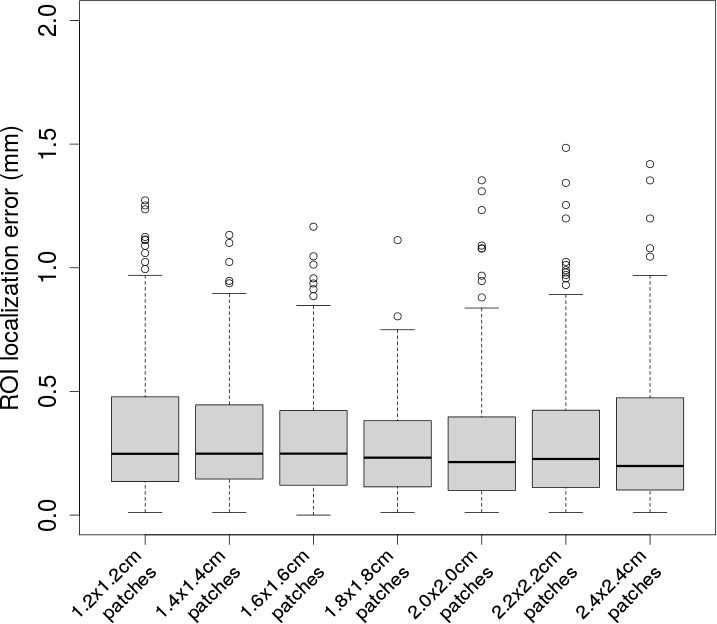}}
\caption{ROI localization error of our system for different sizes of patches. The results are obtained using leave-one-patient-out cross-validation based on the training subjects. In our analyses, we measure the localization error as the Euclidean distance between the estimated ROI location and the one provided by the expert.  As can be seen,  the use of $1.8\times1.8$~cm patches achieves the most stable performance, yielding low   ROI localization error with only a few outliers.}
\label{fig:boxplot_roi_cv}
\end{figure}

\begin{figure}[h]
\centering
\subfloat{\includegraphics[width=1.0\linewidth]{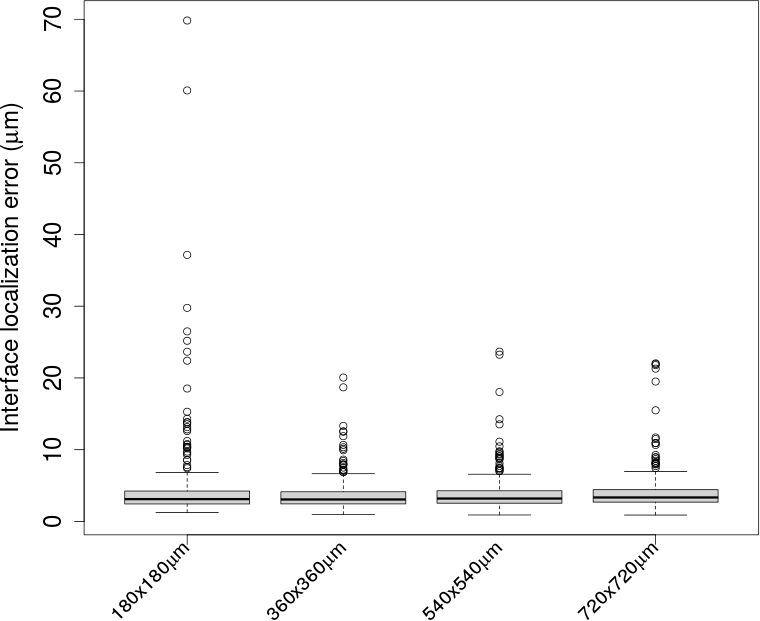}}
\caption{Combined interface localization error for different sizes of patches. The results are produced through a leave-one-patient-out  cross-validation study based on the training subjects. Each box plots show the combined localization error of lumen-intima and media-adventitia interfaces for a different size of patches.  In our analyses, we determine the localization error as the average of absolute vertical distances between our detected boundaries and the expert-annotated boundaries for the interfaces.  As can be seen, while our system shows a high degree of robustness against different sizes of input patches, the use of patches of size  $360\times360$ \SI{}{\micro\metre} achieves slightly lower localization error and fewer outliers. Furthermore, this choice of patches yields higher computational efficiency compared to the larger counterpart patches.}
\label{fig:boxplot_IMT_cv}
\end{figure}

\clearpage

\subsection*{Agreement Analysis}
\begin{figure}[h]
\centering
\subfloat{\includegraphics[width=.92\linewidth]{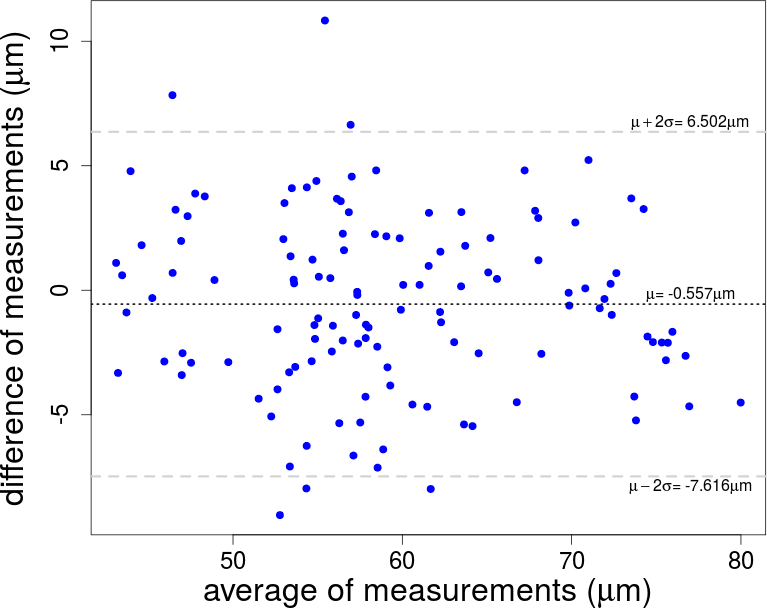}}
\caption{The Bland-Altman plot shows high agreement between our system and the expert for the assessment of intima-media thickness. Each circle in this plot represents a pair of thickness measurements from our method and the expert for a test ROI. In this plot, we have a total of 126 circles corresponding to 44 test videos.}
\label{fig:blandAltman_te_imt_jimmy}
\end{figure}

We further analyze agreement between our system and the expert for the assessment of intima-media thickness. To this end, we use the Bland-Altman plot, which is a well-established technique to measure agreement between different observers. We have shown the Bland-Altman plot for the test subjects in \figurename~\ref{fig:blandAltman_te_imt_jimmy}, where each circle represents a pair of thickness measurements, one from our method and one from the expert. As seen, the majority of circles fall within 2 standard deviations  from the mean error, which suggests a large agreement between the automatically computed thickness measurements and those of the expert.  Furthermore, Pearson product-moment correlation coefficient for the average and difference measurements is -0.097, indicating that the agreement between our method and the expert does not depend on intima-media thickness.


\end{document}